%% file: main.tex
\documentclass[10pt,twocolumn,letterpaper]{article}

\usepackage[pagenumbers]{iccv} 

\usepackage[abs]{overpic}
\usepackage{multirow} 
\usepackage{makecell}
\usepackage{stfloats}
\usepackage{microtype}
\usepackage{enumitem,kantlipsum}
\usepackage{amsmath}
\usepackage{pifont}
\usepackage{colortbl,tabularx}
\usepackage[most]{tcolorbox}
\usepackage{graphicx}
\usepackage{pgfplots}
\usepackage{pifont}
\pgfplotsset{compat=1.17}  
\usepgfplotslibrary{groupplots}  

\newcommand{\dataset}{\textsc{Mapillary Vistas Validation for Fine-Grained Traffic Signs}}

\input{preamble}

\definecolor{iccvblue}{rgb}{0.21,0.49,0.74}
\usepackage[pagebackref,breaklinks,colorlinks,allcolors=iccvblue]{hyperref}

\def\paperID{19} 
\def\confName{ICCV}
\def\confYear{2025}

\title{\dataset: A Benchmark Revealing Vision-Language Model Limitations}

\author{
Sparsh Garg, Abhishek Aich \\
NEC Laboratories, America
}

\begin{document}
\thispagestyle{empty}
\pagestyle{empty}
\maketitle
\input{sec/0_abstract}    
\input{sec/1_intro}
\input{sec/2_related}
\input{sec/3_dataset}
\input{sec/4_methods}
\input{sec/5_results}
\input{sec/6_discussion}

{
    \small
    \bibliographystyle{ieeenat_fullname}
    \bibliography{main}
}

\end{document}

%% file: preamble.tex
\usepackage{ragged2e}
\usepackage{pifont,newunicodechar}
\DeclareUnicodeCharacter{2717}{\ensuremath{\times}}

\DeclareUnicodeCharacter{2717}{\ensuremath{\times}}

\DeclareUnicodeCharacter{2713}{\checkmark}

%% file: sec/0_abstract.tex
\begin{abstract}
Obtaining high-quality fine-grained annotations for traffic signs is critical for accurate and safe decision-making in autonomous driving. Widely used datasets, such as Mapillary, often provide only coarse-grained labels—without distinguishing semantically important types such as \textit{stop signs} or \textit{speed limit signs}.
To this end, we present a new validation set for traffic signs derived from the Mapillary dataset called Mapillary Vistas Validation for Traffic Signs (\textbf{MVV}), where we decompose composite traffic signs into granular, semantically meaningful categories. The dataset includes pixel-level instance masks and has been manually annotated by expert annotators to ensure label fidelity. 
Further, we benchmark several state-of-the-art VLMs against the self-supervised DINOv2 model on this dataset and show that DINOv2 consistently outperforms all VLM baselines—not only on traffic
sign recognition, but also on heavily represented categories
like vehicles and humans. Our analysis reveals significant limitations in current vision-language models for fine-grained
visual understanding and establishes DINOv2 as a strong baseline for dense semantic matching in autonomous driving
scenarios. This dataset and evaluation framework pave the
way for more reliable, interpretable, and scalable perception
systems.

\noindent\textbf{Code and data are available at:} \url{https://github.com/nec-labs-ma/relabeling}
\end{abstract}

%% file: sec/1_intro.tex
\section{Introduction}

Autonomous driving systems rely heavily on large-scale, diverse, and precisely annotated datasets to develop robust perception and decision-making capabilities. A critical aspect of this is the accurate detection and classification of fine-grained traffic sign categories—such as \textit{stop}, \textit{speed limit}, and \textit{yield}—which are essential for ensuring safety and adherence to real-world driving regulations. This task is particularly challenging due to the inherently small size of traffic signs, their distant placement, and their frequent occlusion in complex urban environments.

While the Mapillary Vistas~\cite{neuhold2017mapillary} dataset provides rich urban scene coverage across global cities, its annotation taxonomy treats traffic signs in a coarse, geometry-centric manner—labeling them by orientation (e.g., \textit{traffic-sign-front}, \textit{traffic-sign-back}) rather than semantic content. Such annotations, while sufficient for scene parsing, fail to support decision-critical perception. Autonomous systems cannot infer appropriate maneuvers without knowing whether a visible sign is a \textit{turn restriction}, \textit{stop}, or \textit{do not enter} sign. This semantic gap limits the utility of existing datasets for training fine-grained, behavior-aware perception modules.

Recent progress in open-source vision-language models (VLMs), such as Gemma-3~\cite{team2025gemma} and InternVL-3~\cite{zhu2025internvl3}, has fueled interest in zero shot recognition pipelines for applications related to autonomous driving. These models promise zero-shot generalization across a wide array of visual categories—ostensibly enabling autonomous vehicles to understand fine-grained semantics without extensive supervision. However, we show that this promise does not hold in practice. Through comprehensive benchmarking, we find that VLMs consistently underperform when tasked with recognizing fine-grained traffic sign categories, as well as small and context-sensitive objects such as pedestrians or vehicles.

To this end, we construct a new 2,000-image validation set based on Mapillary Vistas~\cite{neuhold2017mapillary} named Mapillary Vistas Validation for Signs(\textbf{MVV}), where traffic signs are relabeled by expert annotators into 11 semantically meaningful categories including others. We pair this dataset with a benchmarking pipeline that leverages DINOv2~\cite{oquab2023dinov2} for dense visual pattern matching. Our experiments reveal that DINOv2~\cite{oquab2023dinov2} significantly outperforms state-of-the-art VLMs across all major object categories—including vehicles, humans, and traffic signs—when fine-grained semantic precision is required. These findings call into question the effectiveness of VLMs for safety-critical autonomous perception and motivate future research in dense, spatially grounded visual understanding.

To summarize, we make the following contributions:

\begin{itemize}
    \item We release an updated validation set based on Mapillary~\cite{neuhold2017mapillary}, containing over 2,000 high-resolution street-view images annotated with 11 fine-grained traffic sign classes including others. A subset of this data is manually validated by experts to ensure evaluation quality.
    \item We benchmark multiple VLMs (e.g., InternVL-3~\cite{zhu2025internvl3}, Gemma-3~\cite{team2025gemma}) against the self-supervised DINOv2~\cite{oquab2023dinov2} model on a dense visual matching task, and show that VLMs consistently struggle with fine-grained category discrimination for traffic signs. In contrast, DINOv2~\cite{oquab2023dinov2} demonstrates superior spatial localization and class separability. Detailed accuracy results across tasks and prompt strategies are shown in Table~\ref{tab:final_results}, highlighting the consistent underperformance of VLMs on fine-grained traffic sign recognition.

\end{itemize}

%% file: sec/2_related.tex
\section{Related Work}

\subsection{Traffic Signs Datasets}

Traffic sign datasets have long supported the development of perception models in autonomous driving. However, many suffer from limited semantic granularity, class imbalance, or contextual coverage. The German Traffic Sign Detection Benchmark (GTSDB)~\cite{Houben-IJCNN-2013} provides only 900 bounding boxes across 43 sign types, which primarily include direction indicators, speed limit signs, and animal warnings. However, it lacks important categories such as \textit{yield} and \textit{parking} signs, and its geographic scope is limited exclusively to German roads, reducing its generalizability to broader real-world scenarios. Its companion dataset, GTSRB~\cite{Stallkamp-IJCNN-2011}, offers over 50K cropped images but lacks full-scene spatial context and is classification-only. 
TT100K~\cite{Zhe_2016_CVPR} enhances geographic diversity by providing 10,000 Chinese street-view images annotated with 221 traffic sign categories. The dataset covers a wide range of sign types, including prohibitory signs (e.g., \textit{no entry}, \textit{speed limit}), instruction/guide signs (e.g., \textit{direction}, \textit{exit}), and warning signs (e.g., \textit{sharp turn}, \textit{pedestrian crossing}). However, only approximately 45 categories have sufficient sample counts for effective training, and the dataset excludes other important semantic classes such as vehicles and pedestrians, which limits its utility in full-scene urban perception tasks.
MTSD~\cite{ertler2020mapillary} is among the largest datasets, with 52K images and 300+ classes. Yet its combination of human and semi-supervised labels introduces annotation noise, and overlapping composite sign categories complicate fine-grained training. CURE-TSR~\cite{Temel2017_NIPSW} introduces robustness to adverse conditions, but its heavy use of synthetic data reduces real-world applicability. Mapillary Vistas~\cite{neuhold2017mapillary} focuses on panoptic scene understanding but labels traffic signs only with orientation-based tags (e.g., \textit{traffic-sign-front}), omitting semantic content essential for safe autonomous decisions. Our work addresses these gaps by introducing a 2,000-image Mapillary Vistas~\cite{neuhold2017mapillary} validation set where coarse \textit{traffic-sign} labels are decomposed into 11 fine-grained semantic categories including others, offering improved utility for benchmarking visual recognition under realistic driving conditions.

\subsection{Vision-Language Models for Visual Matching}

Vision-language models (VLMs) have shown strong potential in aligning visual content with rich textual semantics, offering a promising direction for reducing manual annotation efforts. Recent models such as InternVL-3~\cite{zhu2025internvl3} and Gemma-3~\cite{team2025gemma} incorporate large-scale vision-text pretraining and multimodal alignment, enabling zero-shot and prompt-based image understanding across open vocabulary categories. In our experiments, we focus on these two models due to their strong performance in real-world driving scenes and their open-source availability.

Despite these advances, current VLMs still exhibit notable limitations when applied to fine-grained relabeling tasks in autonomous driving. Specifically, they often fail to recognize small, distant, or occluded objects such as traffic signs and pedestrian symbols---categories critical for safety in driving systems. Additionally, obtaining accurate predictions frequently requires extensive prompt engineering and iterative query tuning, which introduces complexity and instability into the pipeline. This dependence on carefully curated prompts is not only time-consuming but also difficult to scale across large datasets with varying visual contexts. Moreover, the large model sizes and inference costs associated with InternVL-3\cite{zhu2025internvl3} and Gemma-3~\cite{team2025gemma} pose practical challenges when deploying these systems at scale, particularly in resource-constrained environments.

\begin{table*}[ht]
\centering
\caption{\textbf{Comparison of traffic sign datasets}. MVV improves over prior work in semantic detail, segmentation (segm.), instance (inst.) masks, and geographic coverage. Geographic (Geo.) diversity is abbreviated as: DE = Germany, CN = China.}
\label{tab:dataset-comparison}
\resizebox{\linewidth}{!}{%
\begin{tabular}{lcccccl}
\toprule
\textbf{Dataset} & \textbf{\#Images} & \textbf{\#Classes} & \textbf{Annotation} & \textbf{Geo.} & \textbf{Segm./Inst.} & \textbf{Key Features} \\
\midrule
GTSDB~\cite{Houben-IJCNN-2013} & $\sim$600 & 43 & BBox & ✗ (DE) & \ding{55} & Low scale, no masks, limited classes \\
GTSRB~\cite{Stallkamp-IJCNN-2011} & 50K (crop) & 43 & Cls. only & ✗ (DE) & \ding{55} & No full-scene context \\
TT100K~\cite{Zhe_2016_CVPR} & 10K & 221* & BBox & ✗ (CN) & \ding{55} & Long-tail, no masks, sign-only \\
CURE-TSR~\cite{Temel2017_NIPSW} & 2M+ & Synth. & Cls. only & ✗ & \ding{55} & Synthetic bias, no masks \\
MTSD~\cite{ertler2020mapillary} & 52K & 300+ & BBox & ✓ (Global) & \ding{55} & Semi-auto labels, no masks \\
\textbf{MVV (Ours)} & 2K & 11 (fine) & BBox + Mask & ✓ (Global) & \ding{51} & Manual refinement, segmentable \\
\bottomrule
\end{tabular}%
}
\vspace{0.3em}
\small *Only ~45 classes are frequent enough for robust training.
\end{table*}

%% file: sec/3_dataset.tex
\section{Proposed Traffic Sign Dataset: The MVV Benchmark}

Existing traffic sign datasets such as GTSDB~\cite{Houben-IJCNN-2013}, TT100K~\cite{Zhe_2016_CVPR}, and MTSD~\cite{ertler2020mapillary} offer valuable benchmarks, but are geographically constrained and primarily focused on bounding box annotations. These datasets often lack broader urban scene context, segmentation masks, and comprehensive object classes such as vehicles and pedestrians. In contrast, the Mapillary Vistas dataset~\cite{neuhold2017mapillary} spans a diverse set of global cities and includes high-resolution segmentation masks across 124 semantic classes, making it a rich foundation for building more generalizable benchmarks.

To this end, we construct a refined validation subset from Mapillary Vistas~\cite{neuhold2017mapillary}, focused on traffic sign understanding called ``Mapillary Vistas Validation for Signs" or the \textbf{MVV} benchmark. While the original Mapillary taxonomy includes only coarse-grained labels such as \textit{traffic-sign-front}, these are insufficient for decision-critical tasks in autonomous driving that require recognizing specific signs like \textit{stop}, \textit{yield}, or \textit{turn restrictions}.

To create the \textbf{MVV} benchmark, we randomly selected 2,000 images from the validation set and relabeled all traffic sign instances using expert annotators into 11 semantically meaningful categories including others. The relabeling process reused the original instance masks provided by Mapillary Vistas~\cite{neuhold2017mapillary}, thus preserving pixel-level segmentation while refining the semantic granularity. The updated label taxonomy includes: \textit{Stop Signs}, \textit{Speed Limit Signs}, \textit{Yield Signs}, \textit{Do Not Enter Signs}, \textit{Crosswalk Signs}, \textit{Parking Signs}, \textit{No Parking Signs}, \textit{Roundabout Signs}, \textit{Turn Restriction Signs}, \textit{Cycle Lane Signs}, and  \textit{Others}.

\begin{figure}[ht]
    \centering
    \includegraphics[width=0.9\linewidth]{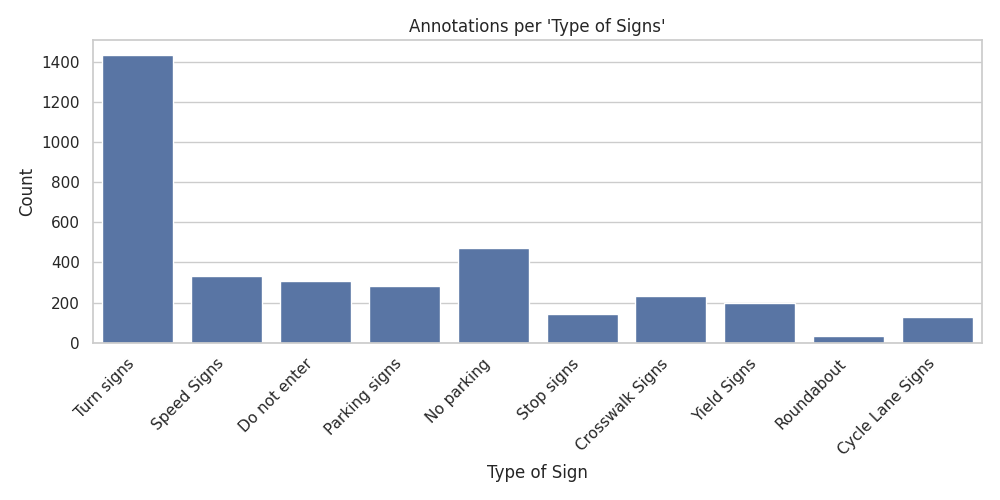}
    \caption{Distribution of annotations per traffic sign category.}
    \label{fig:category-distribution}
\end{figure}

\begin{figure}[ht]
    \centering
    \includegraphics[width=0.9\linewidth]{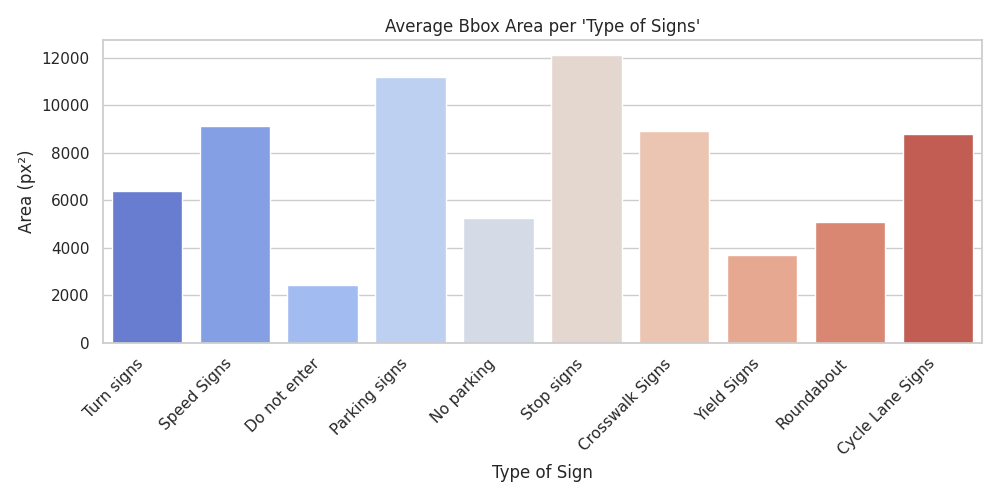}
    \caption{Average bounding box area per category (in pixels).}
    \label{fig:bbox-area}
\end{figure}

\begin{figure}[ht]
    \centering
    \includegraphics[width=0.9\linewidth]{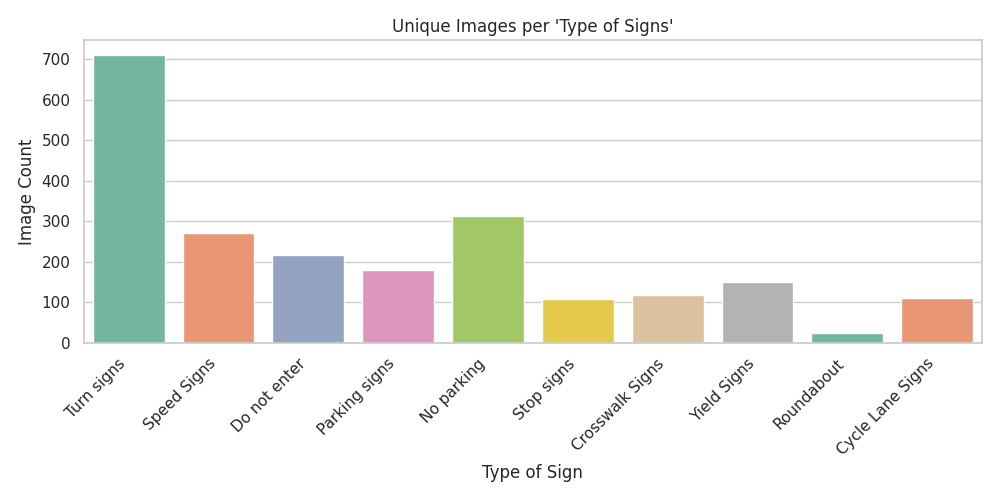}
    \caption{Number of images containing at least one instance of each sign type.}
    \label{fig:image-count}
\end{figure}

This dataset supports both object detection and instance segmentation tasks. Figure~\ref{fig:category-distribution} presents the distribution of annotations per category, illustrating typical urban class imbalance. Figure~\ref{fig:bbox-area} shows the average bounding box area by sign type, and Figure~\ref{fig:image-count} reports category-level image occurrence. Our proposed benchmark enables fine-grained evaluation under realistic driving scenarios, including occlusion, truncation, and scale variation.

%% file: sec/4_methods.tex
\section{VLM Analysis on the MVV Benchmark and Common Objects}
In addition to evaluating VLMs performance on the MVV benchmark, we define a set of relabeling tasks on two widely used autonomous driving datasets: Mapillary Vistas\cite{neuhold2017mapillary} and 	BDD100K\cite{yu2020bdd100k}. These tasks are designed to assess whether VLMs can reliably recognize common, high-frequency object categories such as \textit{vehicles} and \textit{humans}. As shown in Table~\ref{tab:annotation_tasks}, we leverage VLMs to refine coarse-grained annotations into fine-grained semantic categories, revealing significant inconsistencies in VLMs predictions even for widely recognized object types.

\begin{table*}[t]
\centering
\caption{Annotation Tasks for Vehicle, Human, and Traffic Sign Classes in Mapillary and BDD Datasets}
\label{tab:annotation_tasks}

\begin{tabularx}{\textwidth}{cccX}
\toprule
\textbf{Tasks-ID} & \textbf{Dataset} & \textbf{Coarse} & \textbf{Fine} \\ \midrule
$\mathcal{T}_1$ & Mapillary~\cite{neuhold2017mapillary} & Vehicles &
        car, bus, truck, bicycle, motorcycle, boat, train \\ \hline
$\mathcal{T}_2$ & Mapillary~\cite{neuhold2017mapillary} & Humans &
        person, bicyclists, motorcyclists, other rider \\ \hline
$\mathcal{T}_3$ & Mapillary~\cite{neuhold2017mapillary} & Traffic Signs &
        stop signs, speed limit signs, yield signs, do not enter, crosswalk signs,
        parking signs, no parking signs, roundabout signs, turn signs,
        cycle lane signs, others \\ \hline
$\mathcal{T}_4$ & BDD~\cite{bdd100k} & Vehicles &
        car, bus, truck, bicycle, motorcycle \\   
\bottomrule
\end{tabularx}
\end{table*}

\subsection{Relabeling Tasks}

\subsubsection{Mapillary Relabeling.}  
The original Mapillary Vistas~\cite{neuhold2017mapillary} dataset annotates broad object categories such as \textit{vehicle}, \textit{human}, and \textit{traffic sign}. We redefine the relabeling task by converting each coarse annotation into one of several fine-grained subclasses:

\begin{itemize}
    \item \textbf{Vehicles:} \textit{car}, \textit{bus}, \textit{truck}, \textit{bicycle}, \textit{motorcycle}, \textit{boat}, \textit{train}
    \item \textbf{Humans:} \textit{person}, \textit{bicyclist}, \textit{motorcyclist}, \textit{other rider}
    \item \textbf{Traffic Signs:} \textit{stop sign}, \textit{speed limit sign}, \textit{yield sign}, \textit{do not enter}, \textit{crosswalk sign}, \textit{parking sign}, \textit{no parking sign}, \textit{roundabout sign}, \textit{turn sign}, \textit{cycle lane sign}, \textit{no parking}, \textit{others}
\end{itemize}

\subsubsection{BDD100K Relabeling.}  
For BDD100K~\cite{bdd100k}, we perform fine-grained relabeling of the \textit{vehicle} category into: \textit{car}, \textit{bus}, \textit{truck}, \textit{bicycle}, \textit{motorcycle}. These tasks simulate real-world requirements where driving agents must distinguish between semantically similar but behaviorally different entities (e.g., a car vs. a bus vs. a trailer).

\subsection{Baseline VLM Setup and Prompt Strategies}
\label{sec:vlm_prompts}

For our baseline experiments, we utilize the full series of variants from two prominent vision-language models: InternVL-3~\cite{zhu2025internvl3} and Gemma-3~\cite{team2025gemma} with their default settings. Our objective is to assess their ability to perform fine-grained classification from coarse-labeled datasets within the context of autonomous driving. To ensure consistency of VLMs, we ran all VLM experiments twice.

\begin{figure*}[ht]
    \centering
    \includegraphics[width=0.85\linewidth]{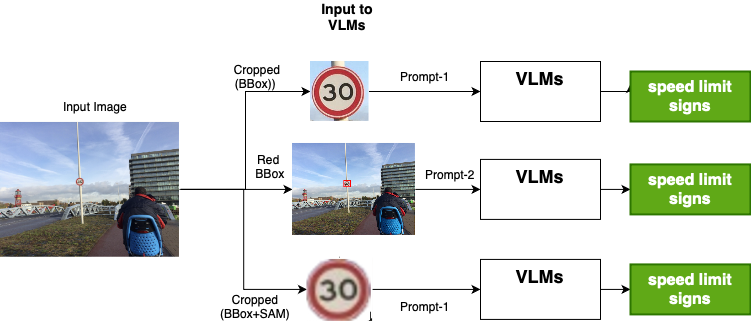}
    \caption{
    \textbf{Prompt-Based Relabeling Pipeline.} Starting from coarse traffic sign annotations, we apply SAM to the full image using mid-point hints and extract fine-grained masks. Cropped inputs using bbox and red bounding box overlays guide the visual prompting strategy. \textit{Prompt 1} queries the cropped image directly, while \textit{Prompt 2} targets a specific red-highlighted region to localize the classification task.
    }
    \label{fig:prompt-flow}
    \vspace{1.5em}
    
    \begin{minipage}[t]{0.45\textwidth}
    \begin{tcolorbox}[colback=gray!5!white, colframe=gray!75!black, title=Prompt 1, fontupper=\small, fontlower=\small]
    \texttt{
        You are an expert in autonomous driving, specializing in analyzing traffic scenes.\\
        Your task is to look at the traffic signs in the image and output the response in the format below.\\
        Think step by step before deciding the answer and return only ONE word answer.\\
        Strictly follow the rules.\\
        \{\\
        Choose ONE category among the following for the most dominant object in the bounding box: <list of objects which VLM needs to classify>\\
    \}}
    \end{tcolorbox}
    \end{minipage}
    \hfill
    \begin{minipage}[t]{0.45\textwidth}
    \begin{tcolorbox}[colback=gray!5!white, colframe=gray!75!black, title=Prompt 2, fontupper=\small, fontlower=\small]
    \texttt{
        You are an expert in autonomous driving, specializing in analyzing traffic scenes.\\
        Your task is to look at the traffic signs in the red bbox and output the response in the format below.\\
        Think step by step before deciding the answer and return only ONE word answer.\\
        Strictly follow the rules.\\
        \{\\
        Choose ONE category among the following for the most dominant object in the bounding box: <list of objects which VLM needs to classify>\\
        \}
    }
    \end{tcolorbox}
    \end{minipage}

\end{figure*}

To this end, we explore three distinct VLM  input with different prompting strategies. also shown in Figure~\ref{fig:prompt-flow}:

\begin{itemize}
    \item \textbf{Cropped Image VLM input:} For each object instance, we crop the region specified by its bounding box and present only the cropped segment to the model. The model is then prompted to classify the object without any access to surrounding scene context.
    
    \item \textbf{Full Image with Red Bounding Box VLM input:} In this setting, the full image is retained, and a red bounding box is drawn around the target object. The model receives the complete scene and is prompted to classify the object within the highlighted region. This setup allows the model to utilize both the object appearance and its contextual surroundings.
    
    \item \textbf{Segmentation-Guided VLM input with SAM~\cite{kirillov2023segment}:} To further improve localization and visual grounding, we apply the \textbf{Segment Anything Model (SAM~\cite{kirillov2023segment})} on the cropped regions to obtain fine object masks. These segmented masks are then used to refine the visual input provided to InternVL~\cite{zhu2025internvl3} and Gemma~\cite{team2025gemma} models. The goal is to assess whether pixel-accurate segmentation improves the model's ability to identify fine-grained categories over bounding-box-based approaches.
\end{itemize}

These strategies allow a controlled comparison of how object isolation, context, and segmentation precision influence the classification accuracy across model variants and driving scene distributions.

\subsection{Language-Free Relabeling Approach}
\begin{figure}[ht]
    \centering
    \includegraphics[width=0.9\linewidth]{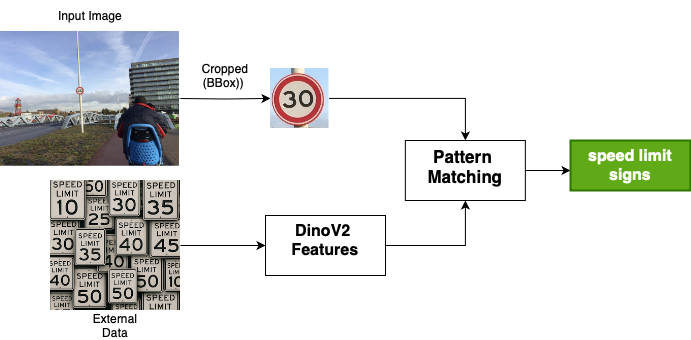}
    \caption{\textbf{Overview of the fine category classification approach using DINOv2}~\cite{oquab2023dinov2} \textbf{feature matching}. Given an input image with a coarse annotation (e.g., "Traffic Sign"), the image is cropped using the coarse bounding box. DINOv2~\cite{oquab2023dinov2} features are extracted from the cropped image and compared with features from open-source fine category samples. The closest match determines the fine-grained label (e.g., "Speed Sign").}
    \label{fig:dinov2}
\end{figure}

 In addition to vision-language models, we also explore a language-free relabeling approach that leverages high-quality visual representations.
\paragraph{Feature-Based Relabeling Using DINOv2.} First, we use visual representations extracted from the self-supervised transformer model \textbf{DINOv2~\cite{oquab2023dinov2}}. Specifically, we compute DINOv2~\cite{oquab2023dinov2} embeddings for cropped object instances drawn from a set of open-source datasets containing fine-grained labels for vehicles, pedestrians, and traffic signs. To ensure unbiased evaluation, we explicitly exclude Mapillary~\cite{neuhold2017mapillary} and BDD~\cite{bdd100k} from both the feature bank and training samples as shown in Figure~\ref{fig:dinov2}. These embeddings are then used within a similarity-based relabeling pipeline: each coarse-labeled object from Mapillary~\cite{neuhold2017mapillary} or BDD~\cite{bdd100k} is matched against the reference feature bank, and its fine-grained category is inferred based on nearest-neighbor retrieval or cosine similarity ranking. This provides an efficient and scalable method to map coarse labels to more specific categories without relying on textual prompts or supervised fine-tuning. Our experiments show that DINOv2~\cite{oquab2023dinov2}-based matching serves as a strong visual-only baseline and offers robust performance, especially in cases where textual ambiguity limits the effectiveness of language-driven models.

\paragraph{Supervised Relabeling with ResNet Classifier.} Second, we also implement a supervised approach by training a ResNet~\cite{he2016deep}-based image classifier with batch size of 32. The classifier is trained on curated subsets of publicly available datasets that contain fine-grained annotations for categories such as vehicles, pedestrians, and traffic signs. Notably, we exclude Mapillary~\cite{neuhold2017mapillary} and BDD~\cite{bdd100k} from the training corpus to ensure unbiased evaluation during relabeling. During inference, we first use the provided coarse category (e.g., \textit{traffic sign}, \textit{vehicle}) to identify the relevant sub-class space, and then pass the cropped object image to the classifier to predict the most likely fine-grained label. This coarse-to-fine classification strategy allows for a controlled label search space, improving reliability and interpretability. This supervised baseline serves as a complementary comparison to both the vision-language and DINOv2~\cite{oquab2023dinov2}-based methods, offering insight into how traditional classification pipelines perform when applied to the relabeling task.

%% file: sec/5_results.tex
\section{Benchmark Evaluations}
As detailed in Table~\ref{tab:annotation_tasks}, we evaluate Vision-Language Models (VLMs) and DINOv2 across four targeted relabeling tasks. Task \texorpdfstring{$\mathcal{T}_1$}{T1} focuses on vehicle classification within the Mapillary Vistas dataset~\cite{neuhold2017mapillary}, refining coarse labels into fine-grained categories such as car, bus, truck, bicycle, motorcycle, boat, trailer, and train. Task \texorpdfstring{$\mathcal{T}_2$}{T2} addresses human categorization in the same dataset, distinguishing between person, motorcyclist, bicyclist, and other riders. Task \texorpdfstring{$\mathcal{T}_3$}{T3} evaluates performance on our MVV traffic sign benchmark, while Task \texorpdfstring{$\mathcal{T}_4$}{T4} targets fine-grained vehicle classification in BDD100K~\cite{bdd100k}, covering car, bus, truck, bicycle, and motorcycle categories.

All VLM experiments were conducted using two NVIDIA A6000 GPUs to ensure consistent evaluation. In contrast, DINOv2 required only a single A6000 GPU. For the supervised classification baseline, training for 50 epochs on a single Tesla GPU required approximately 6 hours. Accuracy was employed as the evaluation metric across all tasks to quantify object recognition performance for both VLMs and DINOv2.

\label{sec:results}

\begin{table*}[htbp]
\centering
\caption{Accuracy comparison across InternVL, Gemma, ResNet-50, ResNet-101 and DINOv2 variants for tasks $\mathcal{T}_1$ to $\mathcal{T}_4$ using different prompts.}
\label{tab:final_results}

\resizebox{0.75\textwidth}{!}{%
\begin{tabular}{lllcccc}
\toprule
\multirow{2}{*}{Model} & \multirow{2}{*}{Variant} & \multirow{2}{*}{Prompt} & \multicolumn{4}{c}{Accuracy} \\
\cmidrule(lr){4-7}
 & & & $\mathcal{T}_1$ & $\mathcal{T}_2$ & $\mathcal{T}_3$ & $\mathcal{T}_4$ \\
\toprule

\multirow{9}{*}{InternVL-3~\cite{zhu2025internvl3}} & \multirow{3}{*}{1B} & Cropped Image             & 0.4394 & 0.0686  & 0.0849   & 0.4972 \\
                          &                     & Red Bounding box          & 0.2888 & 0.0323   & 0.0396   & 0.3253 \\
                          &                     & Cropped Image\_SAM        & 0.2772 & 0.1067   & 0.0921   & 0.4159 \\ \addlinespace

                          & \multirow{3}{*}{2B} & Cropped Image             & 0.4971 & 0.3621  & 0.3231   & 0.2557 \\
                          &                     & Red Bounding box          & 0.3291 & 0.2587  & 0.1703   & 0.2752 \\
                          &                     & Cropped Image\_SAM        & 0.4136  & 0.3551  & 0.3062   & 0.292 \\ \addlinespace

                          & \multirow{3}{*}{8B} & Cropped Image             & 0.6822 & 0.3496   & 0.453   & 0.3933 \\
                          &                     & Red Bounding box          & 0.7681 & 0.4229   &   0.2502   &  0.6171\\
                          &                     & Cropped Image\_SAM        & 0.5932 & 0.2987  & 0.440   & 0.3562 \\ \addlinespace

                          & \multirow{3}{*}{9B} & Cropped Image             & 0.6838  & 0.3931  & 0.46789   & 0.4008 \\
                          &                     & Red Bounding box          & 0.685  & 0.2894  & 0.2825   & 0.7405 \\
                          &                     & Cropped Image\_SAM        & 0.5885 & 0.3659  & 0.45395  & 0.3648 \\ \addlinespace

                          & \multirow{3}{*}{14B}& Cropped Image             & 0.6681 & 0.3576  & 0.4075   & 0.409 \\
                          &                     & Red Bounding box          & 0.6764 & 0.4082 & 0.2873    & 0.7183 \\
                          &                     & Cropped Image\_SAM        & 0.5651 & 0.3155  & 0.3836   & 0.4031 \\ \addlinespace
\midrule
\multirow{6}{*}{Gemma-3~\cite{team2025gemma}}    & \multirow{3}{*}{4B} & Cropped Image             & 0.804  & 0.9044  & 0.2762   & 0.9372 \\
                          &                     & Red Bounding box          & 0.7085  & 0.8768  & 0.16355   & 0.8737 \\
                          &                     & Cropped Image\_SAM        & 0.7551 & 0.8819  & 0.2521   & 0.9368 \\ \addlinespace

                          & \multirow{3}{*}{12B}& Cropped Image             & 0.7926  & 0.9051   & 0.4278   & 0.9285 \\
                          &                     & Red Bounding box          & 0.6641   & 0.7449   & 0.275    & 0.7336 \\
                          &                     & Cropped Image\_SAM        & 0.7061  & 0.8803   & 0.4214   & 0.8909 \\ \addlinespace
\midrule
Classifier               & ResNet-50           & Cropped Image                        & 0.8198 & 0.8847  & 0.2745   & 0.721 \\
Classifier               & ResNet-101           & Cropped Image                        & 0.8826 & 0.8586  & 0.3099   & 0.901 \\
DINO-v2~\cite{oquab2023dinov2}                  & --                  & Cropped Image                        & \textbf{0.935} & \textbf{0.906} & \textbf{0.484} & \textbf{0.95} \\

\bottomrule
\end{tabular}%
}
\end{table*}

\subsection{\texorpdfstring{$\mathcal{T}_1$}{T1}: Relabeling Mapillary Vehicles}
As shown in Table~\ref{tab:final_results}, we report classification results on Mapillary~\cite{neuhold2017mapillary} vehicle categories, which have been subdivided into specific classes: \textit{car}, \textit{bus}, \textit{truck}, \textit{bicycle}, \textit{motorcycle}, \textit{boat}, \textit{trailer}, and \textit{train}. Various InternVL-3~\cite{zhu2025internvl3} and Gemma-3~\cite{team2025gemma} variants were evaluated. We observe a general trend where performance improves proportionally with model size (i.e., number of parameters). However, when compared to DINOv2~\cite{oquab2023dinov2}—where category features were extracted using external data from Object365~\cite{object365}—DINOv2 consistently outperformed all VLM variants.
In terms of efficiency, there is a notable disparity in inference latency: while DINOv2~\cite{oquab2023dinov2} completes feature extraction and classification in a matter of hours, vision-language models require 3–4 days to process 2,000 images due to their sequential text-image processing pipeline.

\subsection{\texorpdfstring{$\mathcal{T}_2$}{T2}: Relabeling Mapillary Humans}
In Table~\ref{tab:final_results}, the \textit{Gemma3-4B} model, when provided with cropped images of humans, demonstrated the ability to reclassify individuals into fine-grained categories such as \textit{person}, \textit{motorcyclist}, \textit{bicyclist}, and \textit{other riders}. Despite this capability, the \textit{DINOv2~\cite{oquab2023dinov2}} model outperformed all vision-language model (VLM) variants, achieving an accuracy of \textbf{0.906}, thereby establishing its superiority in fine-grained human category recognition.
To enhance the fine-grained classification performance of DINOv2~\cite{oquab2023dinov2}, we incorporated external datasets for specific human categories, including
\textit{pedestrains}~\cite{cyclist-detector-training-data-v3_dataset}, \textit{person}~\cite{person-vthiu_dataset}, \textit{bicyclist}~\cite{bike-detect-ct_dataset}, and \textit{motorcyclist}~\cite{helmeteeeeeeeee_dataset}, sourced from large-scale annotated collections such as Robolow. These datasets were used to extract representative visual features that served as exemplars for image pattern matching within the DINOv2 feature space.

\subsection{\texorpdfstring{$\mathcal{T}_3$}{T3}: Relabeling Mapillary Signs}
In Table~\ref{tab:final_results}, the \textit{InternVL-9B} model, when provided with cropped images, demonstrated the capability to relabel traffic signs into fine-grained categories.
Despite this classification ability, InternVL-9B achieved an accuracy of \textbf{0.467}. In contrast, \textit{DINOv2~\cite{oquab2023dinov2}} outperformed all VLM baselines by achieving a slightly higher accuracy of \textbf{0.484}. To extract features for DINOv2~\cite{oquab2023dinov2}, we again leveraged open-source datasets via Roboflow. To increase category diversity and improve model robustness, we sourced traffic sign data from both \textit{U.S.}~\cite{traffic-signs-detection-dpnpl_dataset} and \textit{European}~\cite{european-road-signs_dataset} and \cite{traffic-signs-detection-europe_dataset}datasets.

\subsection{\texorpdfstring{$\mathcal{T}_4$}{T4}: Relabeling BDD Vehicles}
In Table~\ref{tab:final_results}, the \textit{InternVL-9B} model, when provided with cropped images, demonstrated the ability to reclassify BDD~\cite{bdd100k} vehicle instances into fine-grained categories, specifically: \textit{car}, \textit{bus}, \textit{truck}, \textit{bicycle}, and \textit{motorcycle}. Among the vision-language models (VLMs), the \textit{Gemma-4B} variant achieved the highest classification accuracy of \textbf{0.937} using cropped image prompts. However, \textit{DINOv2~\cite{oquab2023dinov2}} again outperformed all VLM variants—including both InternVL and Gemma models—achieving a superior accuracy of \textbf{0.95}. The features used for DINOv2~\cite{oquab2023dinov2} in this experiment were sourced from the same Object365 dataset~\cite{object365} referenced in Table~\ref{tab:final_results}, providing a consistent foundation for category-level vehicle recognition.

%% file: sec/6_discussion.tex
\section{Conclusion}
\label{sec:discussion}
In this work, we revisited the problem of fine-grained semantic recognition for traffic signs in autonomous driving, where precise categorization is essential for safe decision-making. With the observation that widely used datasets like Mapillary Vistas~\cite{neuhold2017mapillary} provide only coarse-grained annotations, we introduced a new human-annotated validation set consisting of 2,000 images derived from Mapillary, where coarse traffic sign categories are decomposed into 11 semantically rich labels including others. Our benchmark supports both detection and segmentation masks, and provides a high-fidelity testbed for evaluating fine-grained recognition in complex driving environments.
Through comprehensive experiments, we showed that state-of-the-art vision-language models (VLMs), despite their zero-shot capabilities, underperform on this task. In contrast, the self-supervised DINOv2 model consistently outperformed all VLM variants across traffic signs, vehicles, and pedestrians—highlighting the limitations of current VLMs in dense, spatially object recognition.
Our findings emphasize the need for spatially attentive and category-discriminative models in safety-critical perception pipelines. We hope our dataset and benchmark will drive further research into scalable, fine-grained recognition for real-world autonomous driving systems.